\documentclass[10pt,twocolumn,letterpaper]{article}

\usepackage{iccv}
\usepackage{times}
\usepackage{epsfig}
\usepackage{graphicx}
\usepackage{amsmath}
\usepackage{amssymb}
\usepackage{bm}
\usepackage{bbm}
\usepackage{booktabs}
\usepackage{multirow}
\usepackage[accsupp]{axessibility}

\usepackage[pagebackref=true,breaklinks=true,letterpaper=true,colorlinks,bookmarks=false]{hyperref}

\iccvfinalcopy 

\def\iccvPaperID{12559} 

\ificcvfinal\fi

\begin{document}

\title{Meta-ZSDETR: Zero-shot DETR with Meta-learning}

\author{
Lu Zhang$^1$ \quad Chenbo Zhang$^1$ \quad  Jiajia Zhao$^2$ \quad Jihong Guan$^3$ \quad Shuigeng Zhou$^1$\thanks{correspondence author}\\
$^1$Shanghai Key Lab of Intelligent Information Processing,  and School of \\ Computer Science,  Fudan University, China\\
$^2$Science and Technology on Complex System Control and \\ Intelligent Agent Cooperation Laboratory, China\\
$^3$Department of Computer Science \& Technology, Tongji University, China\\
{\tt\small \{l\_zhang19,sgzhou\}@fudan.edu.cn,
	cbzhang21@m.fudan.edu.cn } \\
{\tt\small zhaojiajia1982@gmail.com, jhguan@tongji.edu.cn}}

%

\maketitle
\ificcvfinal\thispagestyle{empty}\fi

\begin{abstract}
Zero-shot object detection aims to localize and recognize objects of unseen classes.
Most of existing works face two problems: the low recall of RPN in unseen classes and the confusion of unseen classes with background.
In this paper, we present the first method that combines DETR and meta-learning to perform zero-shot object detection, named Meta-ZSDETR, where model training is formalized as an individual episode based meta-learning task.
Different from Faster R-CNN based methods that firstly generate class-agnostic proposals, and then classify them with visual-semantic alignment module, Meta-ZSDETR directly predict class-specific boxes with class-specific queries and further filter them with the predicted accuracy from classification head.
The model is optimized with meta-contrastive learning, which contains a regression head to generate the coordinates of class-specific boxes, a classification head to predict the accuracy of generated boxes, and a contrastive head that utilizes the proposed contrastive-reconstruction loss to further separate different classes in visual space.
We conduct extensive experiments on two benchmark datasets MS COCO and PASCAL VOC. Experimental results show that our method outperforms the existing ZSD methods by a large margin.
\end{abstract}

\section{Introduction}
Object detection~\cite{FasterRCNN:NIPS} is one of the most fundamental tasks in computer vision.
Most existing object detection methods require huge amounts of annotated training data, which is expensive and time-consuming to acquire.
Meanwhile, in reality novel categories constantly emerge, and there is seriously lack or even nonexistent of visual data of those novel categories for model training, such as endangered species in the wild.
The above issues motivates the investigation of zero-shot object detection, which aims to localize and recognize objects of unseen classes.
\begin{figure}[t]
	\begin{center}
		\includegraphics[width=1.0\linewidth]{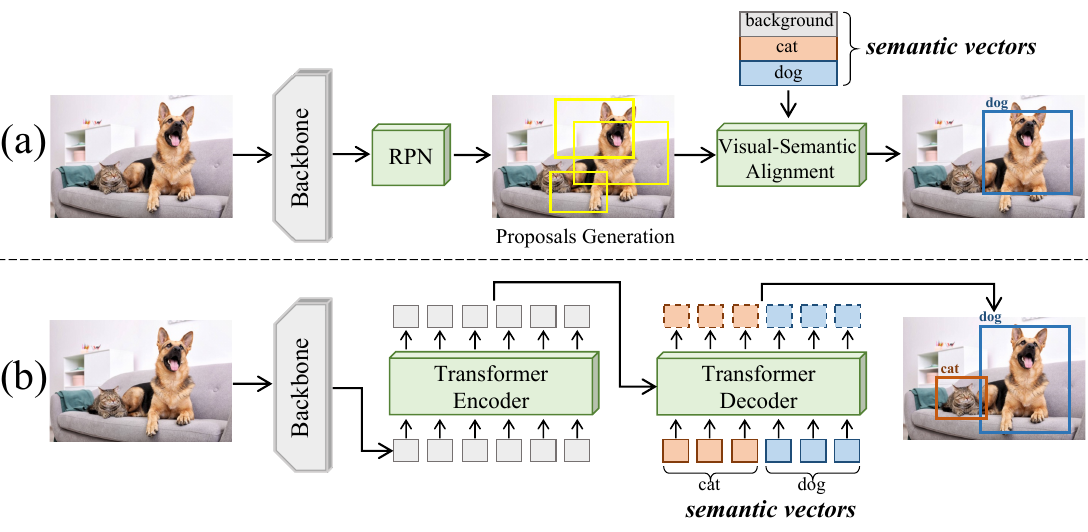}
	\end{center}
	\caption{Zero-shot object detection: Faster R-CNN based methods vs. Meta-ZSDETR. (a): Faster R-CNN based methods firstly generate class-agnostic proposals, and then classify them with different visual-semantic alignment modules. (b): Meta-ZSDETR directly predict class-specific boxes with class-specific queries and further filter them with classification head.}
	\label{fig:introduction}
\end{figure}

A mainstream framework of the existing works that are based on Faster R-CNN, is illustrated in Fig.~\ref{fig:introduction}(a), where the RPN remains unchanged and the RoI classification head is replaced with different visual-semantic alignment modules, such as mapping to the same embedding space to calculate similarity between proposals and semantic vectors~\cite{ZSD_ECCV2018,Semantics-guided-contrastive,graph_propagation_zsd,ZSD_Joint,Background-learnable,multi-space-zsd,zsd-attributes,Simple-zsd,hybrid_region},
synthesizing visual features from semantic vectors~\cite{Robust_region_synthesizer,synthesizing_unseen,Gtnet,zsd_syn_donnot_look,zsd_discriminative_gan} etc.

However, we observe that the existing methods are suboptimal, due to their obvious inherent shortcomings:
\romannumeral1) The proposals from RPN are often not reliable enough to cover all unseen classes objects in an image because of lacking training data, which has also been identified by a recent study~\cite{RPN_overfitting}.
\romannumeral2)
The confusion between background and unseen classes is an intractable problem. Although many previous works have tried to tackle it~\cite{ZSI,synthesizing_unseen,ZSD_ECCV2018}, the results are still unsatisfactory.


Recently, object detection frameworks based on the Transformer  have gained widespread popularity, such as DETR~\cite{DETR}, Deformable DETR~\cite{Deformable-DETR}, etc.
Such architectures are RPN-free and background-free, i.e., they do not involve RPN and background class,
which are naturally conducive to building zero-shot object detection methods.
However, how to build a ZSD method based on DETR detectors poses new challenges.
An intuitive idea is to replace DETR's classification head with a zero-shot classifier based on cosine similarity~\cite{zsd_transformer}. However, such a method simply treats DETR as a large RPN for proposals generation, the overall framework is essentially the same as previous works.

In this paper, we present the \textit{first} method that fully explores DETR detectors and meta-learning to perform zero-shot object detection, named \textbf{Meta-ZSDETR},
which can solve the two problems mentioned above that have plagued the field of ZSD for many years, 
and achieves the state-of-the-art performance.
The comparison of Meta-ZSDETR with previous methods is shown in Fig.~\ref{fig:introduction}.
Different from the previous works that
firstly generate class-agnostic proposals and then classify them with visual-semantic alignment module,
our method utilizes semantic vectors to guide both proposal generation and classification, which greatly improves the recall of unseen classes.
Meanwhile, there is no background class in DETR detectors, which means the confusion between background and unseen classes is no more existent.

In order to detect unseen classes, we formalize the training process as an individual episode based meta-learning task.
In each episode, we randomly sample an image $I$ and a set of classes $\mathcal{C}_{\pi}$, which contains the positive classes that appear in $I$ and negative classes that do not appear.
The meta-learning task is to make the model learn to detect all positive classes of $\mathcal{C}_{\pi}$ on image $I$.
Through the meta-learning task, the training and testing can be unified, i.e., in the model testing, we need only to employ the unseen classes as the set $\mathcal{C}_{\pi}$.
To enable the model to detect an arbitrary class set, we
firstly fuse each object query with a projected semantic vector from the class set $\mathcal{C}_{\pi}$, which transfers the query from class-agnostic to class-specific.
Then, the decoder takes the class-specific query as input and predicts the locations of class-specific boxes, together with the probabilities that the boxes belong to the fused class.
To achieve the above goal, we propose meta-contrastive learning, where all predictions are split into three different types and different combinations of them are chosen to optimize three different heads, i.e., the regression head to generate the locations of class-specific boxes, the classification head to predict the accuracy of generated boxes, and the contrastive head to separate different classes in visual space for performance improving with a contrastive-reconstruction loss.
The bipartite matching and loss calculation are performed in a class-by-class manner, and the final loss is averaged over all classes in the sampled class set $\mathcal{C}_{\pi}$.


In summary, our major contributions are as follows:
\begin{itemize}
	\item We present the first method that explores DETR and meta-learning to perform zero-shot object detection, which formalizes the training as an individual episode based meta-learning task and ingeniously tackles the two problems that plague ZSD for years.
	
	\item  We propose to train the decoder to directly predict class-specific boxes with class-specific queries as input, under the supervision of our meta-contrastive learning that contains three different heads.
	
	\item We conduct extensive experiments on two benchmark datasets MSCOCO and PASCAL VOC to evaluate the proposed method Meta-ZSDETR. Experimental results show that our method outperforms the existing ZSD methods.
	
\end{itemize}

\section{Related work}
\label{sec:related_work}
\subsection{Zero-shot learning}

Zero-shot learning~(ZSL) aims to classify images of unseen classes that do not appear during training.
There are two main streams in ZSL: embedding based methods and generative based methods.
The key idea of embedding based methods is to learn an embedding function that maps the semantic vectors and visual features into the same embedding space, where the visual features and semantic vectors can be compared directly~\cite{akata2013label,bucher2016improving,fu2014transductive,kodirov2017semantic,sung2018learning,zhao2022towards}.
Generative based methods aim to synthesize unseen visual features with variational autoencoder~\cite{kingma2013auto} and generative adversarial networks~\cite{xian2019f}, which convert the ZSL into a fully supervised way~\cite{chen2021free,han2020learning,schonfeld2019generalized,xian2018feature}.


\subsection{Zero-shot object detection}
Zero-shot object detection~(ZSD) has received a great deal of research interest in recent years.
Most of ZSD methods are built on Faster R-CNN~\cite{girshick2015fast}, YOLO~\cite{redmon2016you} and RetinaNet~\cite{lin2017focal,recent_FSOD}.
The process of these methods can be summarized as:
generating class-agnostic proposals and classifying proposals into seen/unseen and background classes.
The main difference of these methods is that different   visual-semantic alignment methods are used to complete the classification of proposals.
These methods can be divided into two categories: 
mapping the semantic vectors and visual features to the same embedding space to calculate similarity~\cite{ZSD_ECCV2018,Semantics-guided-contrastive,graph_propagation_zsd,ZSD_Joint,Background-learnable,multi-space-zsd,zsd-attributes,Simple-zsd,hybrid_region} and
synthesizing visual features from semantic vectors~\cite{Robust_region_synthesizer,synthesizing_unseen,Gtnet,zsd_syn_donnot_look,zsd_discriminative_gan}.
Although previous works have paid great efforts, there are some problems that still have no satisfactory solution, such as the low recall of class-agnostic RPN for unseen classes and the confusion between background and unseen classes. 
These problems may be caused by the incompatibility of ZSD task and proposals-based architecture such as Faster R-CNN.

Different from previous works, Meta-ZSDETR is the first work built on Deformable DETR with meta-learning, where the semantic vectors are guided for class-specific boxes generation, instead of class-agnostic proposals in previous works, resulting in a higher unseen recall and precision. Meanwhile, since there is no background class in DETR detectors, the confusion between background and unseen classes is non-existent.


\section{Method}

\subsection{Problem definition}

Zero-shot object detection~(ZSD) aims to detect objects of unseen classes with model trained on the seen classes.
Formally, the class space $\mathcal{C}$ in ZSD is divided into seen classes $\mathcal{C}^{s}$ and unseen classes $\mathcal{C}^{u}$, where $\mathcal{C} = \mathcal{C}^s \cup \mathcal{C}^u $ and $ \mathcal{C}^{s} \cap \mathcal{C}^{u} = \varnothing$. 
The training set contains objects of seen classes, where each image $I$ is provided with ground-truth class labels and bounding boxes coordinates.
While the test set may contain only unseen objects~(ZSD setting) or both seen and unseen classes~(GZSD setting).
During the training and testing, the semantic vectors $\mathcal{W} = \{\mathcal{W}^{s},\mathcal{W}^{u}\}$ is provided for both seen and unseen classes.



\begin{figure*}[t]
	\begin{center}
		\includegraphics[width=0.9\linewidth]{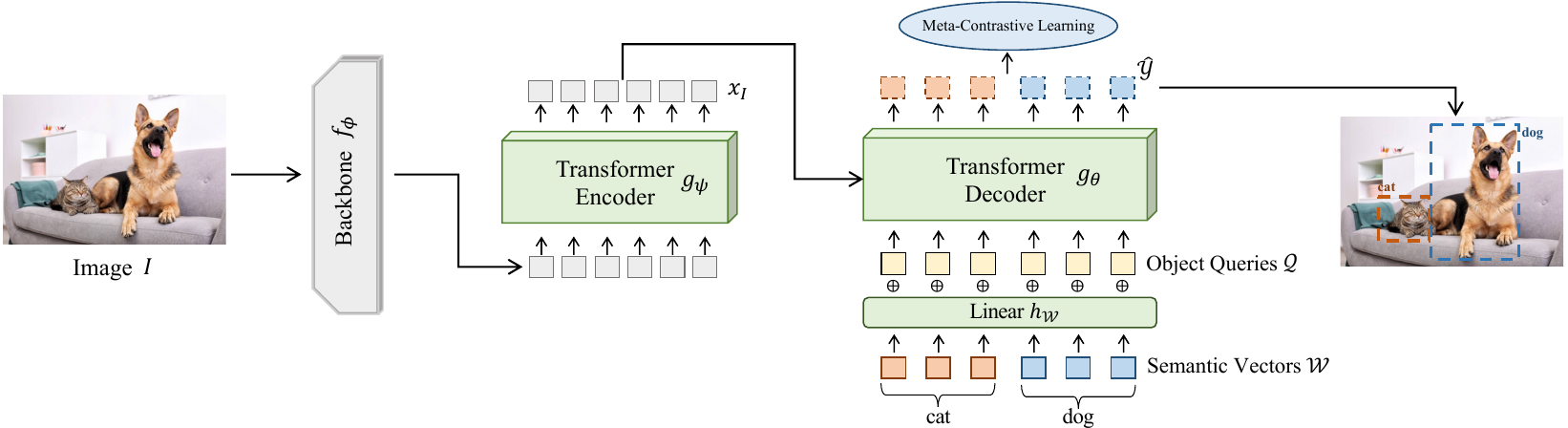}
	\end{center}
	\caption{The framework of Meta-ZSDETR.
	In each episode, a class set $\mathcal{C}_{\pi}$ and an image $I$ are sampled. The meta-learning task is to make the model learn to detect all appeared classes in $\mathcal{C}_{\pi}$. 
	Firstly, the image feature $x_I$ is extracted with backbone and encoder. Then, the projected semantic vectors are added to object queries, making them class-specific. 
	Finally, the decoder $g_{\theta}$ will takes the queries as input and directly predict class-specific boxes.
	To achieve the this, we train our model with proposed meta-contrastive learning.}
	\label{fig:framework}
\end{figure*}

\subsection{Revisit standard DETR in object detection}
To begin with, we review the pipeline of a standard DETR in generic object detection, which contains the following steps:
set prediction, optimal bipartite matching and loss  calculation.
\subsubsection{Set prediction}
For an image $I$, the global representation $x_I$ is extracted by the backbone $f_{\phi}$ and Transformer encoder $g_{\psi}$ successively, which can be expressed as:
\begin{equation}
	\label{eq:feature_extract}
	x_I = g_{\psi}(f_{\phi}(I))
\end{equation}

Then, the decoder  $g_{\theta}$ infers $N$ object predictions $\hat{\mathcal{Y}}$, where $N$ is determined by the number of object queries $\mathcal{Q}$ that serve as learnable positional embedding:
\begin{equation}
	\hat{\mathcal{Y}} = g_{\theta}(x_I,\mathcal{Q})
\end{equation}
where $\hat{\mathcal{Y}} = \{(\hat{c}_i,\hat{b}_i) \}_{i=1}^N$ and $\mathcal{Q} = \{q_i\}_{i=1}^{N}$. For each object query $q_i$, the decoder $g_{\theta}$ will output a prediction box, 
which contains two parts: the predicted class $\hat{c}_i$ and predicted box location $\hat{b}_i$.

\subsubsection{Optimal bipartite matching}


The optimal bipartite matching is to find the minimal-cost matching between the predictions $\hat{\mathcal{Y}} = \{(\hat{c}_i,\hat{b}_i) \}_{i=1}^N$ and ground-truth boxes $\mathcal{Y} = \{(c_i,b_i)\}_{i=1}^N$ (padded with no object $\varnothing$). Therefore, we search for a permutation of $N$ elements $\sigma  \in \mathfrak{S}_{N}$ with lowest cost:
\begin{equation}
	\label{eq:search_permutation}
	\hat{\sigma} = \mathop{\arg\min}_{\sigma  \in \mathfrak{S}_{N}} \sum_{i=1}^{N} \left[ \mathcal{L}_{cls}(c_i, \hat{c}_{\sigma_{i}}) 
	+ \mathcal{L}_{loc}(b_i, \hat{b}_{\sigma_{i}}) \right]
\end{equation}
where $\mathcal{L}_{cls}(c_i, \hat{c}_{\sigma_{i}})$  and $\mathcal{L}_{loc}(b_i, \hat{b}_{\sigma_{i}})$ are matching cost for class prediction and box location with index $\sigma_{i}$, respectively.
Bipartite matching produces one-to-one assignments, where each prediction $(\hat{c}_i,\hat{b}_i)$ is assigned to either a ground-truth box $(c_i,b_i)$ or $\varnothing$ (no object). The permutation for lowest cost is calculated with Hungarian algorithm.

\subsubsection{Hungarian loss}
Hungarian loss is a widely used loss function in DETR, which takes the following form:
\begin{equation}
	\label{eq:loss_Hungarian}
	\mathcal{L}_{Hug} =  \sum_{i=1}^{N} \left[ \mathcal{L}_{cls}(c_i, \hat{c}_{\hat{\sigma}_{i}}) 
	+ \mathbbm{1}_{\{c_i \ne \varnothing \}}  \mathcal{L}_{loc}(b_i, \hat{b}_{\hat{\sigma}_{i}}) \right]
\end{equation}
where $\hat{\sigma}$ is the optimal assignment computed in Eq.(\ref{eq:search_permutation}). 
$\mathcal{L}_{cls}$ is the loss for classification, which usually takes the form of focal loss~\cite{lin2017focal} or cross-entropy loss. 
$\mathcal{L}_{loc}$ is the location loss and usually contains  $l_1$ loss and GIoU loss~\cite{GIOU}.  

\textbf{Challenge:}
Since the standard DETR can only locate the boxes and predict the classes of objects in training set, it is unable to detect unseen classes.
In this paper, we utilize the meta-learning to make the model learn to detect objects according to the inputed semantic vectors, so that the model has the ability to detect objects of any category, as long as the semantic vector of the corresponding category is input.


\subsection{Framework}
We present the framework of Meta-ZSDETR in Fig.~\ref{fig:framework}, which is based on Deformable DETR.
Meta-ZSDETR follows the paradigm of meta-learning. 
The training is performed by episode based meta-learning task.
In each episode, we randomly sample an image $I$ and a class set $\mathcal{C}_{\pi}$. The meta-learning task of each episode is to make the model learn to detect all appeared classes in  $\mathcal{C}_{\pi}$ on image $I$.
Specifically, the image feature is firstly extracted by backbone and Transformer encoder as in Eq.(\ref{eq:feature_extract}).
In order for the decoder to detect categories in $\mathcal{C}_{\pi}$, we add the projected semantic vectors of classes $\mathcal{C}_{\pi}$ to the object queries, making the queries class-specific.
Then, the decoder takes the queries as input and predicts the class-specific boxes directly.
To achieve this, the model is optimized with meta-contrastive learning, which contains a regression head to generate the coordinates of class-specific boxes, a classification to predict the accuracy of generated boxes and a contrastive head that utilize the proposed contrastive-reconstruction loss to further separate different classes in visual space. 

\subsection{Meta-ZSDETR with class-specific queries}
%


To enable the model to detect any unseen class, we fuse the object queries with class semantic information, and make the model learn to predict the bounding boxes for the fused classes.
Such a process is carried out in each meta-learning task.



Specifically, in each episode, we randomly sample a class set $\mathcal{C}_{\pi}$ and an image $I$, 
where the $\mathcal{C}_{\pi}$ satisfies 
$\mathcal{C}_{\pi} \subseteq \mathcal{C}^s$ and each element is unique. 
Meanwhile, 
$\mathcal{C}_{\pi} = \mathcal{C}^{+}_{\pi} \cup \mathcal{C}^{-}_{\pi}$, where $ \mathcal{C}^{+}_{\pi}$ is the positive classes that appeared in image $I$ and $\mathcal{C}^{-}_{\pi}$ is the randomly sampled negative classes that do not appear in $I$ for contrast.
Meanwhile, we denote the size of $\mathcal{C}_{\pi}$ as $L(\mathcal{C}_{\pi})$, where $L(\cdot)$ is the operation that calculate size.
The positive rate $\frac{L(\mathcal{C}_{\pi}^+)}{L(\mathcal{C}_{\pi})}$ is set to $\lambda_{\pi}$, which is a hyperparameter.

Then, the corresponding semantic vectors $\mathcal{W}_{\pi}$ of class set $\mathcal{C}_{\pi}$ are projected from semantic space to the visual space with a linear layer $h_{\mathcal{W}}$:

\begin{equation}
	\widetilde{\mathcal{W}_{\pi}} = h_{\mathcal{W}}(\mathcal{W}_{\pi})
\end{equation}
where $\widetilde{\mathcal{W}_{\pi}}$ is the projected semantic vectors of class set $\mathcal{C}_{\pi}$.
Since $L(\widetilde{\mathcal{W}_{\pi}}) \ll N$, i.e. the number of semantic vectors is smaller than the number of object queries $\mathcal{Q}$, we expand $\widetilde{\mathcal{W}_{\pi}}$ by duplicating each element in $\widetilde{\mathcal{W}_{\pi}}$ for $T$ times,
which satisfies $L(\widetilde{\mathcal{W}_{\pi}})\cdot T \geqslant N$ and  $L(\widetilde{\mathcal{W}_{\pi}})\cdot (T-1) < N$. For redundant elements more than $N$, we drop them.


Then, the projected semantic vectors $\widetilde{\mathcal{W}_{\pi}}$ is added to object queries $\mathcal{Q}$ as follows:
\begin{equation}
	\mathcal{Q}_{\pi} = \mathcal{Q}  \oplus \widetilde{\mathcal{W}_{\pi}}
\end{equation}
where $\mathcal{Q}_{\pi} = \{q^{\pi}_i\}_{i=1}^{N}$ is the class-specific queries that will be inputed into the Transformer decoder $g_{\theta}$ with image feature $x_I$ to generate predictions:
\begin{equation}
	\label{eq:decoder_zsd}
	\hat{\mathcal{Y}} = g_{\theta}(x_I,\mathcal{Q}_{\pi})
\end{equation}
$\hat{\mathcal{Y}} = \{(\hat{\delta}_i,\hat{b}_i) \}_{i=1}^N$ is the set of  predictions, where $\hat{b}_i$ is predicted box location generated with query $q^{\pi}_i$ and $\hat{\delta}_i$ is the probability of box $\hat{b}_i$ belonging to the fused class, i.e. the class of semantic vector that fused to query $q^{\pi}_i$.
Meanwhile, different from the standard DETR that the classification head determines the class of predicted boxes, the class of predicted box $\hat{b}_i$ in Meta-ZSDETR is class-specific and is determined by the class of corresponding query  and 
$\hat{\delta}_i$ only has one dimension to represent the 
probability of $\hat{b}_i$ belongs to the fused class.


\subsection{Meta-contrastive learning}

%


In order for the regression head of decoder $g_{\theta}$ to generate more accurate class-specific box coordinate $\hat{b}_i$  and classification head to have a stronger discriminative ability of further judging the location accuracy of generated $\hat{b}_i$, 
we propose the meta-contrastive learning to train the heads of decoder $g_{\theta}$, i.e. a regression head to generate class-specific boxes, a classification head to filter inaccurate boxes, and moreover a contrastive head to further separate different classes in visual space, which will improve the performance of both seen classes and unseen classes.



\begin{figure}[t]
	\begin{center}
		\includegraphics[width=1.0\linewidth]{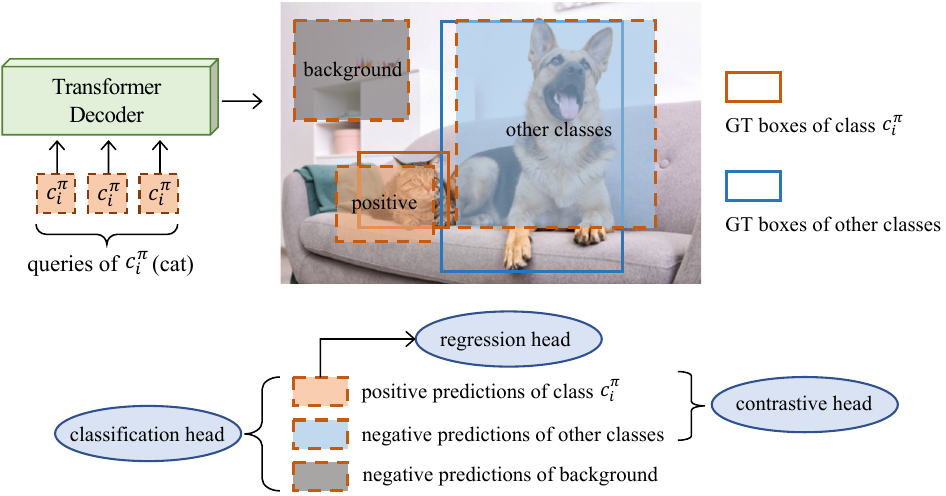}
	\end{center}
	\caption{All predictions of class $c_j^{\pi}$ are split into three different types and different combinations of them are utilized to train three different heads.}
	\label{fig:matching}
\end{figure}

Meta-contrastive learning performs the matching and optimization in a class-by-class manner. 
As shown in Fig.~\ref{fig:matching},
for each class $c_j^{\pi} \in \mathcal{C}_{\pi}$, the decoder takes queries of class  $c_j^{\pi}$ as input, and 
the predictions are split into three types:
1) The positive predictions that are assigned to GT box of class $c_j^{\pi}$ by class-specific bipartite matching. 
2) The negative predictions that are assigned to any other classes than $c_j^{\pi}$.
3) The negative predictions that belong to  background.

Then, we takes different combinations of three types of predictions to train three heads:
a) For the classification head, we take all predictions for the training and give them corresponding positive/negative class-specific targets for class $c_j^{\pi}$.
b) For the regression head, since the regression head aims to generate class-specific boxes of class $c_j^{\pi}$, we only utilize the positive predictions for optimization.
c) For the contrastive head, since our intention is to separate different classes in visual space,  we only use positive predictions of class $c_j^{\pi}$ and negative predictions of other classes for contrastive-reconstruction loss.

\subsubsection{Class-specific bipartite matching}

Our meta-contrastive learning performs the matching in a class-by-class manner. 
For class $c_j^{\pi} \in \mathcal{C}_{\pi}$, the predictions generated by queries of class $c_j^{\pi}$ are selected for bipartite matching, which is denoted as 
$\hat{\mathcal{Y}}_{c_j^{\pi}} = \{(\hat{\delta}_{\tau_i},\hat{b}_{\tau_i}) \}_{i=1}^{T_{\tau}}$, 
where $\{\tau_i\}_{i=1}^{T_{\tau}}$ are the indexes for queries of class $c_j^{\pi}$ in $N$ queries. Since each query is duplicated for $T$ or $T-1$ times, we denote the length uniformly as $T_{\tau}$.

Then, the ground-truth matching targets are revised for each class $c_j^{\pi} \in \mathcal{C}_{\pi}$.
The class-specific matching labels are denoted as $\Delta_{c_j^{\pi}} = \{\delta_i^{c_j^{\pi}}\}_{i=1}^{T_{\tau}}$ (padded with no objects $\varnothing$), which satisfy:
\begin{equation}
	\delta_i^{c_j^{\pi}} = 
	\left\{\begin{matrix}
		1, \quad   c_i = c_j^{\pi} \\
		0, \quad   c_i \neq c_j^{\pi}
	\end{matrix}\right.
\end{equation}
where $c_i$ is the origin class label for ground-truth box $b_i$. The revised matching labels are generated according to whether $c_i$ is equal to $c_j^{\pi}$.
Then, the  matching targets $\mathcal{Y}_{c_j^{\pi}} = \{(\delta_i^{c_j^{\pi}}, b_i)\}_{i=1}^{T_{\tau}}$ is utilized for bipartite matching of class $c_j^{\pi}$, where a permutation of $T_{\tau}$ elements $\sigma  \in \mathfrak{S}_{T_{\tau}}$ with lowest cost is search as:
\begin{equation}
	\label{eq:search_permutation_c}
	\hat{\sigma} = \mathop{\arg\min}_{\sigma  \in \mathfrak{S}_{T_{\tau}}} \sum_{i=1}^{T_{\tau}} \left[ \mathcal{L}_{cls}(\delta_i^{c_j^{\pi}}, \hat{\delta}_{\tau_{\sigma_{i}}} )
	+ \mathcal{L}_{loc}(b_i, \hat{b}_{\tau_{\sigma_{i}}}) \right]
\end{equation}
where $\hat{\sigma}$ is the optimal assignment
between  matching targets $\mathcal{Y}_{c_j^{\pi}}$ 
and predictions $\hat{\mathcal{Y}}_{c_j^{\pi}}$. $\mathcal{L}_{cls}$ and  $\mathcal{L}_{loc}$ are the same as that in Eq.(\ref{eq:search_permutation}).

\subsubsection{Loss function}

Based on  $\hat{\sigma}$, 
predictions in $\hat{\mathcal{Y}}_{c_j^{\pi}}$ are split into three types mentioned above, i.e. positive and two types of negative ones, and loss function of class $c_j^{\pi}$ for three heads are calculated as follows:
\begin{small}
\begin{equation}
\begin{aligned}
	\label{eq:loss_c}
	\mathcal{L}_{c_j^{\pi}} = \sum_{i=1}^{T_{\tau}} \Big[ \mathcal{L}_{cls}(\delta_i^{c_j^{\pi}}, \hat{\delta}_{\tau_{\hat{\sigma}_{i}}} )
	&+ \mathbbm{1}_{(c_i = c_j^{\pi})} \mathcal{L}_{loc}(b_i, \hat{b}_{\tau_{\hat{\sigma}_{i}}}) \Big]  
	+ \mathcal{L}_{cont}
\end{aligned}
\end{equation}
\end{small}
where the loss $\mathcal{L}_{c_j^{\pi}}$ for class $c_j^{\pi}$ is composed of classification loss $\mathcal{L}_{cls}$, regression loss $\mathcal{L}_{loc}$ and contrastive-reconstruction loss $\mathcal{L}_{cont}$.

\textbf{Classification loss.} 
$\mathcal{L}_{cls}$ takes all predictions as input and makes the model learn to distinguish whether a predicted box belongs to  $c_j^{\pi}$. The predicted box has the label $\delta_i^{c_j^{\pi}} = 1$ if and only if it is assigned to a ground-truth box with class label $c_j^{\pi}$.
We implement $\mathcal{L}_{cls}$ with focal loss~\cite{lin2017focal}.

\textbf{Regression loss.}
Since our intention is to generate class-specific boxes, i.e. input decoder with query of class $c_j^{\pi}$ and output box of class $c_j^{\pi}$, 
we only select the ground-truth box with class label $c_j^{\pi}$ for optimization, i.e. $\delta_i^{c_j^{\pi}} = 1$, making the predicted boxes to be closer to GT boxes with class label $c_j^{\pi}$. 
The $\mathcal{L}_{loc}$ is implement with $l_1$ loss and GIoU loss~\cite{GIOU}.

\textbf{Contrastive-reconstruction loss.}
In ZSD tasks, the original visual space is not well-structured due to the lack of discriminative information, many previous works solved it by introducing a reconstruction loss to guide the distribution of visual features. Here, we combine the reconstruction loss and contrastive loss~\cite{he2020momentum,Zhang_2021_CVPR,hifsod} to bring a higher intra-class compactness and inter-class separability of the visual structure. 

In particular, we project the last hidden features of decoder to the semantic space, where the projected hidden features of positive predictions are constraint to be as close as possible to $\omega_j^{\pi}$, which is the semantic vector of class $c_j^{\pi}$, and the negative ones are constraint to be as far as possible to $\omega_j^{\pi}$.
Formally, we denote the last hidden feature of optimal box $(\hat{\delta}_{\tau_{\hat{\sigma}_{i}}}, \hat{b}_{\tau_{\hat{\sigma}_{i}}})$ that matched to GT box $(c_i,b_i)$ as $z_{\tau_{\hat{\sigma}_{k}}}$ and 
the $\mathcal{L}_{cont}$ is formulated as:

\begin{equation}
	\mathcal{L}_{cont} = \frac{1}{N_{pos}} \sum_{i=1}^{T_{\tau}} 
	\mathbbm{1}_{(c_i = c_j^{\pi})} \mathcal{L}_{cont}({z_{\tau_{\hat{\sigma}_{i}}}})
\end{equation}
\begin{equation}
\label{eq:infoNCE}
\mathcal{L}_{cont}({z_{\tau_{\hat{\sigma}_{i}}}}) = -\log \frac{\exp[h_{\rho}(z_{\tau_{\hat{\sigma}_{i}}}) \cdot \omega_j^{\pi} / \kappa]}
 { \sum_{k=1}^{T_{\tau}} \mathbbm{1}_{(c_k \neq \varnothing)} \exp[h_{\rho}(z_{\tau_{\hat{\sigma}_{k}}}) \cdot \omega_j^{\pi} / \kappa]
 }
\end{equation}
where $h_{\rho}$ is a linear layer that project the hidden feature to the semantic space. 
$N_{pos}$ is the number of positive predictions of class $c_j^{\pi}$.
$\kappa$ is a temperature hyper-parameter as in InfoNCE~\cite{infoNCE}.
The optimization of the above loss function increases the
instance-level similarity between projected hidden features of positive predictions with semantic vector $\omega_j^{\pi}$ and space the negative ones. 
As a result, visual features of the same class will form a tighter cluster.

\textbf{Total loss function.}
We compute the loss function with Eq.(\ref{eq:loss_c}) for each class $c_j^{\pi} \in \mathcal{C}_{\pi}$, separately.
For negative classes $\mathcal{C}^{-}_{\pi}$, only classification loss is calculated.
Finally, the loss of current episode is averaged over all classes in sampled class set $\mathcal{C}_{\pi}$:
\begin{equation}
	\mathcal{L} = \frac{1}{L(\mathcal{C}_{\pi})} \sum_{j=1}^{L(\mathcal{C}_{\pi})} \mathcal{L}_{c_j^{\pi}}
\end{equation}
where $L(\mathcal{C}_{\pi})$ is the number of classes in $\mathcal{C}_{\pi}$. We utilize $\mathcal{L}$ for model optimization.



\section{Experiments}

\subsection{Datasets and splits}
Following previous works~\cite{Semantics-guided-contrastive,Robust_region_synthesizer}, we takes two benchmark datasets for evaluation: PASCAL VOC 2007+2012~\cite{voc} and MS COCO 2014~\cite{coco}.

\textbf{Datasets:}
PASCAL VOC contains 20 classes of objects for object detection. More specifically, the PASCAL VOC 2,007 is composed of 2,501 training images, 2,510 validation images, and 5,011 test images. The PASCAL VOC 2012 dataset comprises 5,717 training images and 5,823 validation images, without test images released.
MS COCO 2014 is a benchmark dataset designed for object detection and semantic segmentation tasks, which contains 82,783 training and 40,504 validation images from 80 categories.
For PASCAL VOC and MS COCO, we adopt the FastText~\cite{mikolov2017advances} to extract the semantic vectors following~\cite{synthesizing_unseen,Robust_region_synthesizer}.

\textbf{Seen/unseen splits:}
All splits follow the previous setting~\cite{Semantics-guided-contrastive,Robust_region_synthesizer}. 
For PASCAL VOC, we use a 16/4 split proposed in ~\cite{hybrid_region}.
For MS COCO, we follow the same procedures described in ~\cite{ZSD_ECCV2018,visual-semantic-alignment,Robust_region_synthesizer} to take 2 different splits, i.e. 48/17 and 65/15.
For all datasets, 
the images that contains unseen classes in the training set are removed to guarantee that unseen objects will not be available during training.

\subsection{Evaluation protocols}
We adopt the widely-used evaluation protocols proposed in~\cite{ZSD_ECCV2018,hybrid_region}.
For PASCAL VOC, mAP with IoU  threshold 0.5 is used to evaluate the performance. 
For MS COCO, mAP and recall@100 with three different IoU threshold~(i.e. 0.4,0.5 and 0.6) are utilized for evaluation.
For GZSD setting that contains both seen and unseen classes, the performance is evaluated by Harmonic Mean~(HM).

\subsection{Implementation details}
We build Meta-ZSDETR on Deformable DETR~\cite{Deformable-DETR} with ResNet-50~\cite{resnet} as backbone. 
The number of queries $N$ is set to 900. 
For the sampled class set $\mathcal{C}_{\pi}$, the positive classes consist of the classes that appear in the image and the negative classes is sampled from $\mathcal{C}^s$.
 The positive rate $\lambda_{\pi}$ is set to 0.5.
The number of Transformer encoder layers and decoder layers is set to 6. 
The temperature hyper-parameter  $\kappa$ in Eq.(\ref{eq:infoNCE}) is set to 0.2.
We train our model for total 500,000 iterations with batch size 16, i.e. each iteration contains 16 episodes in parallel.
Following Deformable DETR, different coefficients are utilized to weight different loss functions, where 1.0 is used for classification loss, 5.0 is used for $l_1$ loss of regression head, 2.0 is used for GIoU loss of regression head and 1.0 is used for contrastive-reconstruction loss.
More details can refer to our code.

\subsection{Comparison with existing methods}

\begin{table}
	\begin{center}
		\renewcommand\tabcolsep{9.0pt}
		\begin{tabular}{lcccc}
			\toprule
			\multirow{2}{*}{Method} & 
			\multirow{2}{*}{ZSD} & 
			\multicolumn{3}{c}{GZSD} \\
		    \cmidrule{3-5}
			& &  S & U & HM \\
			\midrule 
			SAN \cite{rahman2018zero}  & 59.1  & 48.0    & 37.0    & 41.8 \\
			HRE \cite{Demirel2018ZeroShotOD}  & 54.2  & {62.4}  & 25.5  & 36.2 \\
			PL \cite{rahman2020improved}   & 62.1  & -     & -     & - \\
			BLC \cite{zheng2020background}  & 55.2  & 58.2  & 22.9  & 32.9 \\
			SU \cite{hayat2020synthesizing}   & 64.9  & -     & -     & - \\
			Robust-Syn \cite{Robust_region_synthesizer} & {65.5}     & 47.1     & {49.1}     & {48.1} \\
			ContrastZSD \cite{Semantics-guided-contrastive}  & 65.7  & 63.2 & 46.5 & 53.6 \\
			Meta-ZSDETR & \textbf{70.3} & \textbf{67.6}  
			& \textbf{56.3} &  \textbf{61.4} \\
			\bottomrule
		\end{tabular}
	\end{center}
	\caption{The results of mAP in PASCAL VOC  with IoU=0.5 under ZSD and GZSD settings. Here, “S” denotes seen classes, “U” denotes unseen classes and “HM” denotes harmonic mean. }
	\label{tab:voc-results}
\end{table}

\begin{table}
	\begin{center}
		\begin{tabular}{lccccc}
			\toprule
			Method & car   & dog   & sofa  & train & mAP \\
			\midrule
			SAN \cite{rahman2018zero}  & 56.2  & 85.3  & {62.6}  & 26.4  & 57.6 \\
			HRE \cite{Demirel2018ZeroShotOD}  & 55.0    & 82.0    & 55.0    & 26.0    & 54.5 \\
			PL \cite{rahman2020improved}   & 63.7  & 87.2  & 53.2  & 44.1  & 62.1 \\
			BLC \cite{zheng2020background}  & 43.7  & 86    & 60.8  & 30.1  & 55.2 \\
			SU \cite{hayat2020synthesizing}   & 59.6  & \textbf{92.7}  & 62.3  & 45.2  & 64.9 \\
			Robust-Syn \cite{Robust_region_synthesizer}   & {60.1}     & {93.0}     & 59.7     & {49.1}     & {65.5} \\
			ContrastZSD \cite{Semantics-guided-contrastive} & 65.5 & 86.4 & 63.1 & 47.9 & 65.7  \\
			Meta-ZSDETR & \textbf{69.0 }& {92.4 }& \textbf{65.7} & \textbf{54.1} & \textbf{70.3} \\
			\bottomrule
		\end{tabular}
	\end{center}
	\caption{Class-wise AP and mAP on unseen classes in PASCAL VOC under ZSD setting.
	}
	\label{tab:voc-class-wise}
\end{table}

\subsubsection{PASCAL VOC}
We present the results of PASCAL VOC in Tab.~\ref{tab:voc-results}, where we can see that our method performs best among all existing methods under both ZSD and GZSD settings, and lift the mAP in PASCAL VOC to a higher level. 

Specifically, in ZSD setting, Meta-ZSDETR achieves 70.3 mAP and outperform the second-best model ContrastZSD~\cite{Semantics-guided-contrastive} by a large margin of 4.6 mAP, which is the first time to boost the performance of ZSD setting on PASCAL VOC to over 70 mAP.

For GZSD setting, our method also achieves SOTAs in all three metrics, i.e. mAP on seen classes, unseen classes and harmonic mean, which brings about 4.4, 9.8 and 7.8 points improvement, respectively.
It is worth noting that the improvement of our method on unseen classes in GZSD setting is extremely large, which  proves that our method  has a strong generalization on unseen classes, and can alleviate the problem that the unseen classes tend to be misclassified into seen classes to a certain extent.
We also report class-wise mAP in ZSD setting in Tab.~\ref{tab:voc-class-wise}, where our method achieves the best performance on 3 classes.

\begin{table}
	\begin{center}
		\renewcommand\tabcolsep{1.1pt}
		\begin{tabular}{lccccc}
		\toprule
		\multirow{2}[4]{*}{Method} & \multirow{2}[4]{*}{Split} & \multicolumn{3}{c}{Recall@100} & mAP \\
		\cmidrule(r){3-5}   \cmidrule(r){6-6}       &       & IoU=0.4   & IoU=0.5   & IoU=0.6   & IoU=0.5 \\
		\midrule
		DSES \cite{bansal2018zero}   & 48/17 & 40.2     & 27.2     & 13.6     & 0.5 \\
		TD \cite{li2019zero}   & 48/17 & 45.5     & 34.3     & 18.1     & - \\
		PL \cite{rahman2020improved}   & 48/17 & -     & 43.5     & -     & 10.1 \\
		BLC \cite{zheng2020background}  & 48/17 & 51.3     & 48.8     & 45.0     & 10.6 \\
		ZSDTR \cite{zsd_transformer} & 48/17 & 51.8 & 48.5 & 44.5 & 10.4  \\
		Robust-Syn \cite{Robust_region_synthesizer}   & 48/17 & {58.1}     & {53.5}     & {47.9}     & {13.4} \\
		ContrastZSD \cite{Semantics-guided-contrastive} & 48/17 & 56.1 & 52.4 & 47.2 & 12.5 \\ 
		Meta-ZSDETR & 48/17 &  \textbf{62.3} & \textbf{59.8} & \textbf{54.2} & \textbf{15.1}  \\
		\midrule
		
		PL \cite{rahman2020improved}   & 65/15  & -     & 37.7     & -     & 12.4 \\
		BLC \cite{zheng2020background}   & 65/15 & 57.2     & 54.7     & 51.2     & 14.7 \\
		SU \cite{hayat2020synthesizing}  & 65/15 & 54.4     & 54.0     & 47.0     & 19.0 \\
		ZSDTR \cite{zsd_transformer} &65/15&  63.8&60.3 &56.5 &13.2 \\
		Robust-Syn \cite{Robust_region_synthesizer}   & 65/15 & {65.3}     & {62.3}     & {55.9}     & {19.8} \\
		ContrastZSD \cite{Semantics-guided-contrastive} &
		65/15 &  62.3 & 59.5 &  55.1 &  18.6  \\
		Meta-ZSDETR &  65/15 & \textbf{69.1} & \textbf{66.7} &  \textbf{59.0} & \textbf{22.5} \\
		\bottomrule
		\end{tabular}
	\end{center}
	\caption{ZSD performance of Recall@100 and mAP with different IoU thresholds on MS COCO dataset.
	}
	\label{tab:coco-zsd}
\end{table}

\begin{table}
	\begin{center}
		\renewcommand\tabcolsep{3pt}
		\begin{tabular}{lccccccc}
		 \toprule
		\multirow{2}[4]{*}{Method} & \multirow{2}[4]{*}{Split} & \multicolumn{3}{c}{Recall@100} & \multicolumn{3}{c}{mAP} \\
		\cmidrule(r){3-5}     \cmidrule(r){6-8}   &       & S     & U     & HM    & S     & U     & HM \\
		\midrule
		PL \cite{rahman2020improved}   & 48/17 & 38.2     & 26.3     & 31.2     & 35.9     & 4.1     & 7.4 \\
		BLC \cite{zheng2020background}  & 48/17 & 57.6     & 46.4     & 51.4     & 42.1     & 4.5     & 8.2 \\
		ZSDTR \cite{zsd_transformer} & 48/17 & 74.3 & 48.4& 60.5&48.5 & 5.6 & 9.5\\
		{\fontsize{9}{10}\selectfont Robust-Syn \cite{Robust_region_synthesizer} }
		& 48/17 & {59.7}     & {58.8}     & {59.2}     & {42.3}     & {13.4}     & {20.4} \\
		{\fontsize{9}{10}\selectfont
		ContrastZSD \cite{Semantics-guided-contrastive}} &
		48/17 &  65.7 & 52.4 & 58.3 & 45.1 & 6.3 & 11.1 \\
		Meta-ZSDETR  & 48/17 & \textbf{74.3} & \textbf{59.0} & \textbf{65.8}  & \textbf{48.7} & \textbf{14.6} &  \textbf{22.5}\\
		\midrule
		PL  \cite{rahman2020improved}  & 65/15 & 36.4     & 37.2     & 36.8     & 34.1     & 12.4     & 18.2 \\
		BLC \cite{zheng2020background}  & 65/15 & 56.4     & 51.7     & 53.9     & 36.0     & 13.1     & 19.2 \\
		SU \cite{hayat2020synthesizing}  & 65/15 & 57.7     & 53.9     & 55.8     & 36.9     & 19.0     & 25.1 \\
		ZSDTR \cite{zsd_transformer} & 65/15 & 
		69.1 & 59.5 & 61.1 & 40.6 & 13.2 & 20.2 
		\\
		{\fontsize{9}{10}\selectfont Robust-Syn \cite{Robust_region_synthesizer} }  & 65/15 & {58.6}    & {61.8}     & {60.2}     & {37.4}     & {19.8}     & {26.0} \\
		{\fontsize{9}{10}\selectfont
		ContrastZSD \cite{Semantics-guided-contrastive} } &
		65/15 & 62.9& 58.6 &60.7 &40.2 &16.5 &23.4      \\
		Meta-ZSDETR & 65/15 & \textbf{71.1} &\textbf{65.4}& \textbf{68.1} &\textbf{45.9} & \textbf{21.7}  & \textbf{29.5} \\
		\bottomrule
		\end{tabular}
	\end{center}
	\caption{GZSD performance of Recall@100 and mAP with IoU=0.5 on MS COCO dataset.
	}
	\label{tab:coco-gzsd}
\end{table}

\subsubsection{MS COCO}

We perform experiments on MS COCO, where the results of ZSD setting is shown in Tab.~\ref{tab:coco-zsd} and the results of GZSD setting is shown in Tab.~\ref{tab:coco-gzsd}. We can see that Meta-ZSDETR achieves the best results in all metrics under all settings.

For ZSD setting, we can see that mAP of our method in 48/17 and 65/15 splits outperforms the second-best by a margin of 1.7 and 2.7 mAP, respectively, which demonstrates that our method generalizes well to unseen classes.
Meanwhile, we can see that Recall@100 decrease as the IoU increases in all methods. Compared with other methods, Meta-ZSDETR has a smaller drop, which is benefit from that our decoder can generate more accurate boxes with class semantic information as input.

For GZSD setting, our method achieves SOTAs in both seen and unseen classes. 
The harmonic means of mAP under 48/17 and 65/15 splits are improved from 20.4 to 22.5, and from 26.0 to 29.5, demonstrating the effectiveness and superiority of our method.
Meanwhile, the Recall@100 also improves due to the powerful class-specific boxes generation capabilities.
We also report the class-wise AP in 65/15 split of MS COCO, which can be found in our supplementary material.


\subsection{Ablation study}
We analyze the effects of various components in Meta-ZSDETR. 
Unless otherwise specified, the experiments are carried out on MS COCO with 65/15 split under GZSD setting and use mAP with IoU=0.5 as metric.

\begin{table}
	\begin{center}
		\begin{tabular}{lcccc}
			\toprule
			$\mathcal{L}_{loc}$ &  $\mathcal{L}_{cls}$ & $\mathcal{L}_{cont}$ & Seen & Unseen \\
			\midrule
			$\checkmark$ & & & 39.9 & 14.5    \\
			$\checkmark$ & $\checkmark$ & & 44.8 & 20.6    \\
			$\checkmark$ &   &  $\checkmark$ & 40.6 & 15.3 \\
			$\checkmark$ & $\checkmark$ & $\checkmark$      &  \textbf{45.9} & \textbf{21.7}  \\
			\bottomrule
		\end{tabular}
	\end{center}
	\caption{Ablation study of different combinations of loss functions.
	}
	\label{tab:ab-loss-all}
\end{table}

\textbf{Effects of different loss functions.}
Here, we analyze the effects of three loss functions in meta-contrastive learning.
We utilize different combinations of regression loss $\mathcal{L}_{loc}$, classification loss $\mathcal{L}_{cls}$ and contrastive-reconstruction loss $\mathcal{L}_{cont}$ to optimize the model, and show the results in Tab.~\ref{tab:ab-loss-all}.
Since the regression loss is necessary, we keep it for all combinations.
For model without classification loss $\mathcal{L}_{cls}$, we do not perform the boxes filter and directly use the class-specific boxes predicted from regression head, where the scores are generated randomly.
As we can see, if the model is only trained with regression head to generate class-specific boxes, it can achieve a mAP of 14.5 in unseen classes, which is relatively low, but also surpasses many previous methods.
Adding the classification loss will greatly boost the performance of unseen classes to 20.6 mAP, which thanks to the powerful discriminative ability of the classification head that can filter inaccurate boxes.
Meanwhile, the contrastive-reconstruction loss can improve the performance with and without the classification head about 1 point.
Finally, the combination of three losses achieves the best performance.

\begin{table}
\begin{center}
	\renewcommand\tabcolsep{3.0pt}
	\begin{tabular}{lccccc}
		\toprule
	 	Heads & $\hat{\mathcal{Y}}_{pos}$ & $\hat{\mathcal{Y}}_{other}$ & $\hat{\mathcal{Y}}_{bg}$ & Seen &  Unseen  \\
		\midrule
		\multirow{3}{*}{Classification Head} & 
		\checkmark & \checkmark      &  & 43.7 & 17.9 \\
		& \checkmark & & \checkmark  & 44.8 & 19.0  \\
		& \checkmark & \checkmark & \checkmark & \textbf{45.9} & \textbf{21.7}\\
		\hline 
		\multirow{2}{*}{Regression Head} & 
		\checkmark   & & &   \textbf{45.9} & \textbf{21.7}  \\
		& \checkmark & \checkmark & & 42.1 & 16.5 \\
		\hline 
		\multirow{3}{*}{Contrastive Head} &
		\checkmark  &  & &   45.1 & 21.0    \\
	&	\checkmark & \checkmark &  & \textbf{45.9} & \textbf{21.7} \\
		& \checkmark & \checkmark & \checkmark & 45.4 & 21.3 \\
		\bottomrule
	\end{tabular}
\end{center}
\caption{Ablation study on using different combinations of predictions to train three heads.
}
\label{tab:ab-head-training}
\end{table}

\textbf{Study for training of different heads.}
As describe,
based on the class-specific bipartite matching for class $c_j^{\pi}$, 
all predictions are split into three different types: the positive predictions $\hat{\mathcal{Y}}_{pos}$ assigned to GT boxes of $c_j^{\pi}$, 
the negative predictions $\hat{\mathcal{Y}}_{other}$ assigned to other classes 
and the negative predictions $\hat{\mathcal{Y}}_{bg}$ that belong to background.
Here, we study different combinations of them to train three heads and the results are shown in Tab.~\ref{tab:ab-head-training}.
We can see that:
1) For classification head, since it aims to filter all kinds of negative predictions, using all predictions to train it can achieve the best performance.
2) For regression head, if we train it with GT boxes of all classes, i.e. using $\hat{\mathcal{Y}}_{pos}$ and $\hat{\mathcal{Y}}_{other}$, the regression head will degenerate into a class-agnostic RPN, which will greatly reduce the recall of unseen classes, thus lead to a lower mAP.
3) For contrastive head, on one hand, if we only use $\hat{\mathcal{Y}}_{pos}$ for training, it will degenerate into a reconstruction loss, which has been widely used in previous works and it will bring a 0.4 mAP improvement in unseen classes compared with the version without it.
On the other hand, compared with using all predictions, 
removing background predictions will make the contrastive head focus on
distinguishing $\hat{\mathcal{Y}}_{other}$ and inputed semantic vectors of class $c_j^{\pi}$, thus brings more improvement.

\begin{figure}[t]
	\begin{center}
		\includegraphics[width=1.0\linewidth]{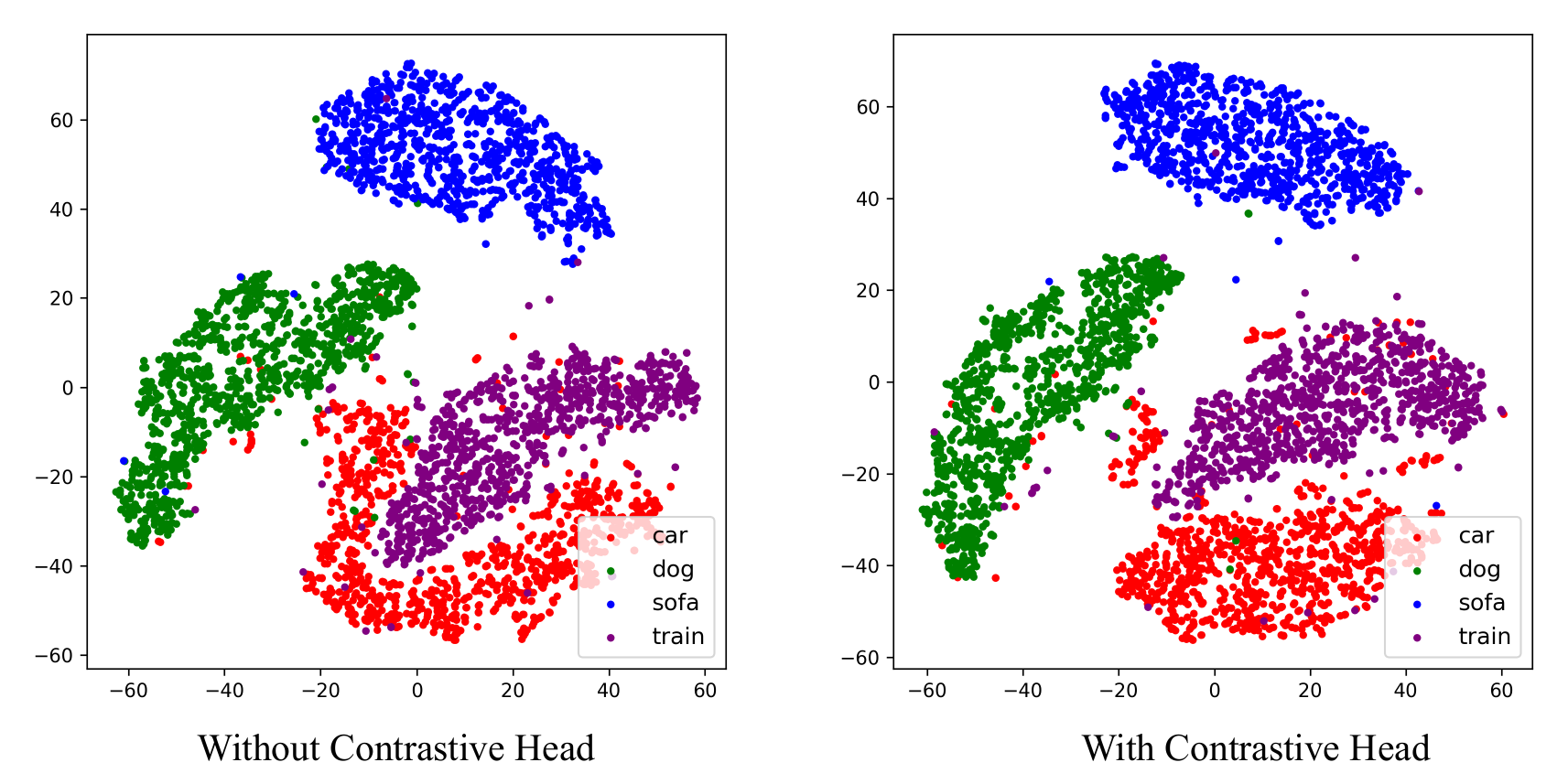}
	\end{center}
	\caption{The t-SNE visualization of the last hidden layer of decoder. We can see that the contrastive head can separate different classes in visual space.}
	\label{fig:vis-tsne}
\end{figure}

\textbf{Visualization for contrastive-reconstruction loss.}
Here, we study the influence of contrastive-reconstruction loss on visual space by visualizing the distribution of hidden features with t-SNE.
We visualize the last hidden features of decoder in unseen classes of PASCAL VOC.
The result is shown in Fig.~\ref{fig:vis-tsne}.
As we can see, our contrastive-reconstruction loss can further separate different classes in visual space and bring a higher intra-class compactness and inter-class separability of the visual structure. 


\begin{figure}[t]
	\begin{center}
		\includegraphics[width=0.8\linewidth]{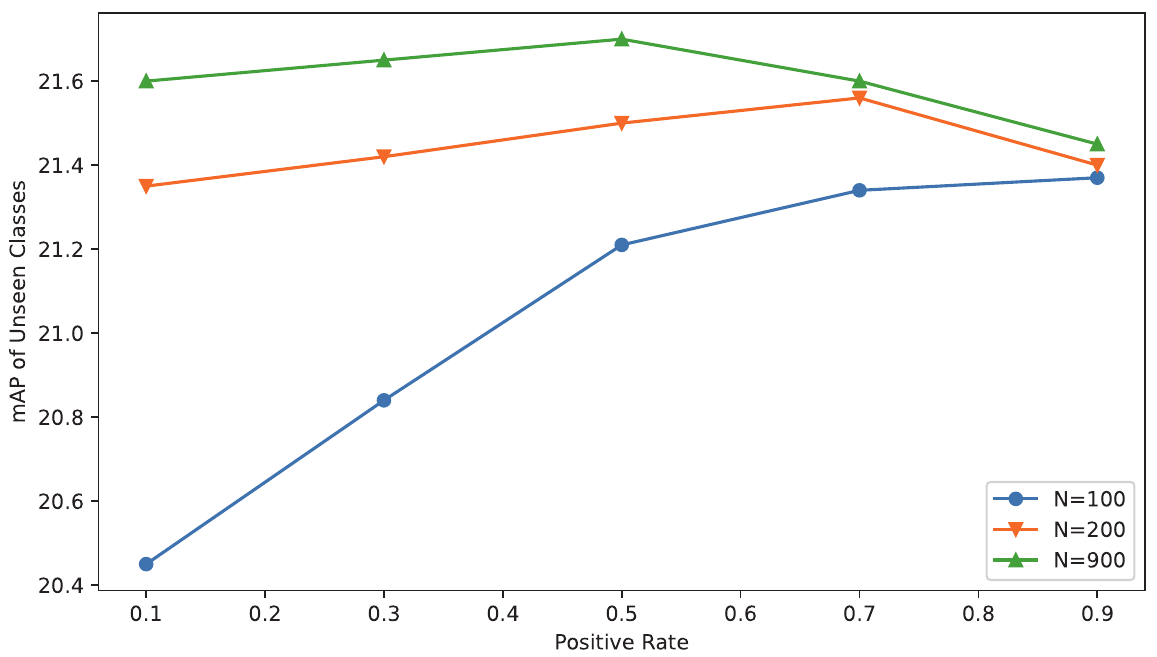}
	\end{center}
	\caption{The effect of number of queries  and positive rate of $\mathcal{C}_{\pi}$}
	\label{fig:positive}
\end{figure}

\textbf{Effect of number of queries and positive rate.}
We study the effect of number of queries $N$ and positive rate $\lambda_{\pi}$ of sampled class set $\mathcal{C}_{\pi}$. 
We found that $\lambda_{\pi}$ have different influence under different number of queries $N$.
We change the positive rate $\lambda_{\pi}$ in different settings of $N$ and report the mAP of converged model in unseen classes. 
In each episode, we control $\lambda_{\pi}$ by sampling different number of negative classes.
All models are trained for 500,000 episodes.
The results are shown in Fig.~\ref{fig:positive}.
As we can see, a larger $N$ tends to have a better performance due to a higher recall, and of course a higher amount of calculation. 
Meanwhile, when $N$ is small (e.g. 100), a small positive rate will greatly reduce the amount of positive queries for training, thereby reducing the model performance.
When $N$ is large (e.g. 900), the number of positive queries is guaranteed and more negative queries are need for the classification head to learn to distinguish among them. Therefore, we can see the best performance is achieved when $\lambda_{\pi}$ is 0.5 and $N$ is 900.

\section{Conclusion}

In this paper, we present the first work that combine DETR and meta-learning to perform zero-shot object detection, which formalize the training as individual episode based meta-learning task. 
In each episode, we randomly sample an image and a class set. The meta-learning task is to make the model learn to detect all appeared classes of the sampled class set on the image.
To achieve this, we train the decoder to directly predict class-specific boxes with class-specific queries as input, under the supervision of our meta-contrastive learning that contains three different heads.
We conduct extensive experiments on the benchmark datasets MSCOCO and PASCAL VOC. Experimental results show that our method outperforms the existing ZSD methods.
In the future, we will focus on further performance improvement.

\textbf{Acknowledgement.}
Jihong Guan was supported by National Natural Science Foundation of China (NSFC) under grant No.~U1936205.

{\small
\bibliographystyle{ieee_fullname}
\bibliography{egbib}
}

\end{document}